%% file: conference.tex
\documentclass[conference]{IEEEtran}
\IEEEoverridecommandlockouts
\usepackage[noadjust]{cite}
\usepackage{amsmath,amssymb,amsfonts}
\usepackage{algorithmic}
\usepackage{graphicx}
\usepackage{textcomp}
\usepackage{xcolor}
\def\BibTeX{{\rm B\kern-.05em{\sc i\kern-.025em b}\kern-.08em
    T\kern-.1667em\lower.7ex\hbox{E}\kern-.125emX}}

\usepackage{mathtools} 
\usepackage{booktabs} 

\usepackage{makecell}
\usepackage{caption}
\usepackage{subcaption}
\usepackage{lscape}
\usepackage{multirow}
\usepackage{bm}
\usepackage{pgfplots}
\pgfplotsset{compat=1.18}

\usepackage[hidelinks]{hyperref}

\usepackage[capitalize,noabbrev]{cleveref}
\Crefname{section}{Section}{Sections}
\crefname{section}{Sec.}{Secs.}
\Crefname{table}{Table}{Tables}
\crefname{table}{Tab.}{Tabs.}
\Crefname{appsec}{Appendix}{Appendices}
\crefname{appsec}{App.}{Apps.}
\Crefname{equation}{Equation}{Equations}
\crefname{equation}{Eq.}{Eqs.}

\usepackage{amsmath,amsfonts,amsthm,dsfont}

\usepackage[normalem]{ulem}

\def\our{\mbox{ProPML}}

\usepackage{pifont}
\usepackage{tcolorbox}

\begin{document}

\title{\our{}: Probability Partial Multi-label Learning\\
}


\author{\IEEEauthorblockN{\L{}ukasz Struski\IEEEauthorrefmark{1}, Adam Pardyl\IEEEauthorrefmark{1}\IEEEauthorrefmark{2}\IEEEauthorrefmark{3}, Jacek Tabor\IEEEauthorrefmark{1} and Bartosz Zieli\'nski\IEEEauthorrefmark{1}\IEEEauthorrefmark{3}}

\IEEEauthorblockA{\IEEEauthorrefmark{1}Jagiellonian University, Faculty of Mathematics and Computer Science \\
Emails: \{lukasz.struski, jacek.tabor, bartosz.zielinski\}@uj.edu.pl, adam.pardyl@doctoral.uj.edu.pl}
\IEEEauthorblockA{\IEEEauthorrefmark{2}Jagiellonian University, Doctoral School of Exact and Natural Sciences}
\IEEEauthorblockA{\IEEEauthorrefmark{3}IDEAS NCBR}

}

\maketitle

\begin{abstract}
Partial Multi-label Learning (PML) is a type of weakly supervised learning where each training instance corresponds to a set of candidate labels, among which only some are true. In this paper, we introduce \our{}, a novel probabilistic approach to this problem that extends the binary cross entropy to the PML setup. In contrast to existing methods, it does not require suboptimal disambiguation and, as such, can be applied to any deep architecture. Furthermore, experiments conducted on artificial and real-world datasets indicate that \our{} outperforms existing approaches, especially for high noise in a candidate set.
\end{abstract}

\section{Introduction}\label{sec:intro}

Deep neural networks are highly effective in many practical applications. However, their success is heavily dependent on the availability of a large dataset with accurate labeling. Obtaining such datasets is challenging due to the cost and discrepancies between labeling experts, like in~\cite{armato2011lung}, where four radiologists review each lung CT independently. Consequently, many contemporary datasets are weakly labeled, forcing researchers to propose adequate learning strategies within the Weakly Supervised Learning (WSL) paradigm~\cite{zhou2018brief}.

Partial Multiple-label Learning (PML)~\cite{xie2021partial,ridnik2021asymmetric,sun2022deep} is a type of WSL problem where each training instance corresponds to a set of candidate labels, among which only some are true (see~\Cref{fig:pll}). It occurs in many real-world tasks, such as sentiment analysis and document categorization~\cite{liu2021emerging} or classification of images, audio and video~\cite{sun2022deep}.

Simple PML approaches treat all candidate labels as relevant and employ the off-the-shelf multi-label image classification methods~\cite{chen2019multi,ridnik2021asymmetric}. However, they perform poorly because of the misleading false labels from the candidate set. Therefore, many methods adopt a disambiguation strategy, which aims to identify the relevant labels in the candidate set, e.g. using prior knowledge~\cite{sun2019partial,zhang2020partial} or auxiliary information~\cite{xie2021partial}. They show improvement compared to simple solutions but have two issues. First, prior knowledge or auxiliary information can be inaccessible in many real-world scenarios. Second, disambiguation is usually defined on the whole dataset, which is not suitable for deep models. As a remedy, the most recent methods~\cite{sun2022deep} introduce curriculum-based disambiguation with consistency regularization. However, they require complex training that alternates between learning the model and updating the disambiguation weights, which results in a suboptimal solution.

\begin{figure}[t]
  \centering
  \begin{tabular}{@{}r@{\quad}l@{}}
    \begin{minipage}{.5\columnwidth}
      \includegraphics[width=\linewidth, height=60mm]{images/FirstFig.pdf}
    \end{minipage}
    &
    \begin{minipage}[c]{.48\columnwidth}
        \begin{tcolorbox}[width=.9\linewidth, halign=center, colframe=blue!50, colback=white, boxsep=1mm, arc=3mm]
          Candidate labels:
          \begin{itemize}
            \setlength{\itemsep}{1pt}
            \setlength{\parskip}{0pt}
            \setlength{\parsep}{0pt}
            \item[\ding{51}] Apple
            \item[\ding{51}] Banana
            \item[] Lemon
            \item[\ding{51}] Lime
            \item[\ding{51}] Orange
            \item[] Peach
            \item[\ding{51}] Pineapple
          \end{itemize}
      \end{tcolorbox}
    \end{minipage}
  \end{tabular}
  \caption{In partial multiple-label learning, each training instance corresponds to a set of candidate labels. Only some of them are true (here, checkmarked), but we do not know which. This situation can appear, e.g., if many experts label the same image. Some of them give correct answers, while others can make mistakes.}
    \label{fig:pll}
\end{figure}

In this paper, we introduce \our{}\footnote{The code of \our{} is provided together with the Supplementary Materials at \url{https://github.com/gmum/propml}} (abbr. from Probabilistic Partial Multiple-label Learning), a novel probabilistic approach to partial multiple-label learning problems that address the above-mentioned issues. For this purpose, we establish a probability function composed of two components. The first encourages the model to find true labels in the candidate set, while the second penalizes it for predicting the labels outside of the candidate set. We show that \our{} can potentially be applied to any deep architecture and target tasks (such as classification, detection, or segmentation). Moreover, the experiments conducted on seven artificial and five real-world datasets indicate that it outperforms existing methods, especially if the number of fake labels in the candidate set increases (see Figure~\ref{fig:voc_plot}).

\input{voc_plot.tex}

Therefore, our contributions can be summarized as follows:
\begin{itemize}
    \item We introduce \our{}, a novel probabilistic approach to the PML problem, which does not require a suboptimal disambiguation strategy or other auxiliary algorithms.
    \item Our approach requires only loss function modification and, as such, can be applied to any deep architecture and any target tasks.
    \item We empirically prove that \our{} outperforms the existing, more composite methods, both for artificial and real-world datasets.
\end{itemize}

\section{Related works}

Partial Multi-label Learning (PML) deals with noisy labels, combining two popular branches of machine learning: multi-label learning and partial-label learning (PLL)~\cite{zhang2017disambiguation,gong2021top,yan2020partial,lv2020progressive,struski2022propall}. In PLL, each training instance corresponds to a set of labels among which only one is true. PML, on the other hand, extends this problem to a multi-label setup, where each training instance corresponds to a set of labels, among which some (one or more) are true.

Multiple methods have been proposed to solve the PML problem. Most of them are based on disambiguation. They try to recover the ground-truth labels from the candidate set and then use them in a standard multi-label setup.
For instance, methods proposed in~\cite{xie2018partial}, PML-FP and PML-LC, that estimate the confidence that labels are true by exploring the structural information of feature or label space.
Another approach proposed in~\cite{sun2019partial} denoises the candidate labels using low-rank and sparse matrix decomposition.
Method proposed in~\cite{zhang2020partial} estimates label confidence using either pairwise label ranking (PAR-VLS) or maximum a posteriori reasoning (PAR-MAP).
More complicated approaches use two-phase disambiguation. PML-LRS~\cite{sun2019partial} splits the observed label set into two label matrices, one with true and one with false labels. Then, the former is constrained to be low rank using the augmented Lagrange multiplier algorithm~\cite{lin2010augmented} to prevent an overfitting method, and the second is constrained to be sparse. PML-NI~\cite{xie2021partial} uses a similar technique. However, it additionally jointly learns the multi-label classifier and noisy label identifier under the supervision of the observed noise-corrupted label matrix.
However, the above methods may suffer from the problem of cumulative error in the propagation process.

More recent methods move toward deep learning, like ASL~\cite{ridnik2021asymmetric} loss function-based approach. It dynamically down weights easy negative samples, hard thresholds (discards) the easiest ones, and removes possibly mislabeled samples. This idea was further extended in CDCR~\cite{sun2022deep} with a curriculum-based disambiguation that progressively identifies ground-truth labels while incorporating the varied difficulties of different classes, such as class imbalance or easiness of distinguishing.
Finally, many works transfer the PML problem to various learning scenarios, including multi-view learning~\cite{chen2020multi}, semi-supervised learning~\cite{xie2020semi}, adversarial training~\cite{yan2021adversarial}.

\section{Probabilistic Partial Multi-label Learning}

The main idea behind our Probability Partial Label Learning (\our{}) is to extend Binary Cross Entropy (BCE) loss to the PML setup as closely as possible. For this purpose, we establish a probability function constructed out of two components. The first one encourages the model to find true labels in the candidate set, while the second one penalizes it for predicting the labels outside of it.

To provide a detailed description of our method, let us recall that in partial multi-label learning, each training sample $x$ corresponds to a candidate set $S \subset \{1, \ldots, C\}$, which contains at least one true label and some amount of false labels ($C$ corresponds to the number of classes). Moreover, let $p_c$, where $c \in C$, be the probability of predicting class c, i.e. obtaining $1$ at $c$th output of any deep architecture. Then, our \our{} loss is defined as
\begin{equation}
\label{eq:promil}
L_{ProPML} =
-\log\sum\limits_{i \in S} p_i - \lambda\cdot\sum\limits_{j \not\in S}\log (1-p_j),
\end{equation}
where $\lambda$ is the hyperparameter responsible for the balance between the accurate prediction of labels from $S$ and the remaining labels, and $\sum_{i \in S} p_i \in [0, |S|]$.

\begin{figure}[t]
    \centering
    \includegraphics[width=\columnwidth]{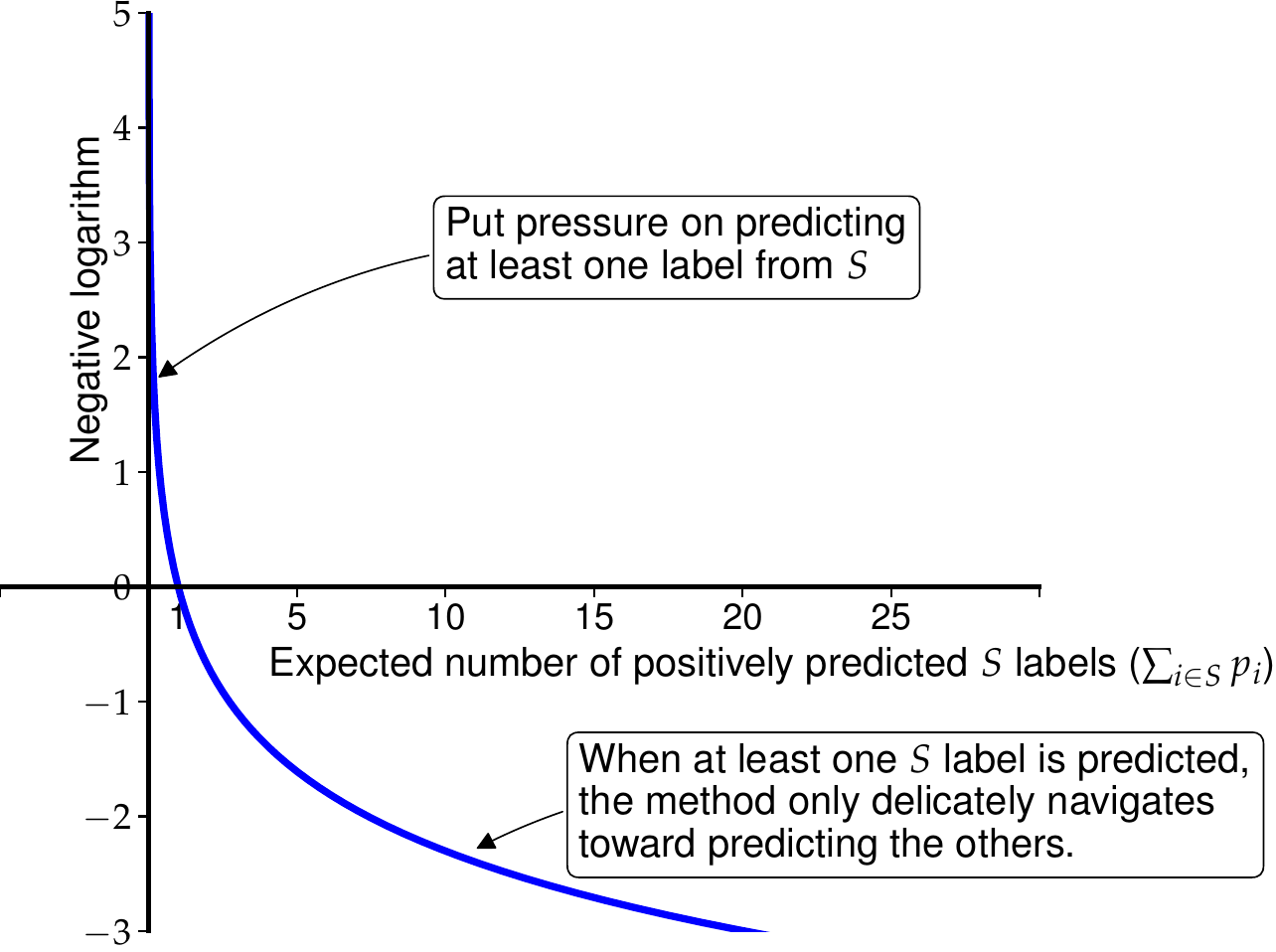}
    \caption{The first component of \our{} loss corresponds to the expected number of positively predicted S labels. If this expected value is small, then none of the labels from $S$ is predicted, and we put strong pressure on predicting at least one of them. However, when the expected is higher than $1$, at least one label from $S$ is probably predicted. Hence, we only delicately navigate toward predicting other S labels, which can be false. We should keep in mind that each $p_i \in [0, 1]$, therefore $\sum_{i \in S} p_i \in [0, |S|]$.}
    \label{fig.promil}
\end{figure}

To understand the idea behind~\our{}, let us analyze its two components separately. The first component $-\log\sum_{i \in S} p_i$ corresponds to the labels from a candidate set, among which only some labels are true. Its value is very high if none of the candidate set labels is predicted. However, as soon as at least one label from $S$ is predicted, the value of this component goes beyond $0$ and decreases more smoothly, together with increasing sum $\sum_{i \in S} p_i$ (see~\Cref{fig.promil}). This way, \our{} puts high pressure on predicting at least one of the candidate labels, and after that, only delicately navigates toward predicting other $S$ labels. From this perspective, \our{} fundamentally differs from the existing PML loss functions, like ASL~\cite{ridnik2021asymmetric}, which aims to predict all labels from a candidate set during the entire training.

The second component $-\sum_{j \not\in S}\log (1-p_j)$ corresponds to labels outside the candidate set, which we know are all false. Therefore, this component is taken directly from BCE loss and, similarly like in BCE, it aims to prevent predicting labels from outside $S$.

Finally, we would like to note that when $S$ is one element set (a single label case) and $\lambda=1$, the \our{} loss reduces to standard BCE loss.

\section{Experiments}

In this section, we evaluate the effectiveness of \our{} by comparing its classification performance with a few existing state-of-the-art PML approaches. We used a variety of metrics to analyze the performance, including average precision, coverage loss, Hamming loss, label ranking, and one error~\cite{zhang2013review}. For the first four metrics, a lower value indicates better performance. For the average precision, a higher value is better. We run the experiments on several real-world and synthetic datasets.

\paragraph{Datasets.}
We started with evaluating our model on five artificial datasets\footnote{\url{http://mulan.sourceforge.net/datasets-mlc.html}} corrupted to the MLP setup: Bibtex, Birds, Enron, Medical, and Scene. To obtain corrupted datasets, we first train the SVM classifier on the original labels, and then we analyze probabilities predicted by this classifier. As false labels, we take those with the highest probability (excluding originally true labels). We consider three scenarios, 50\%, 100\%, or 150\%,  depending on the ratio of false labels to true labels. Furthermore, we use five real-world PML datasets: MIRFlickr~\cite{zhang2020partial}, Music-emotion~\cite{huiskes2008mir}, Music-style~\cite{huiskes2008mir}, YeastBP, and YeastCC~\cite{yu2018feature}. Finally, we use two vision datasets, the PASCAL Visual Object Classes Challenge 2017 (VOC2007)~\cite{everingham2009pascal} and Microsoft Common object in Context 2014 (COCO2014)~\cite{lin2014microsoft}. Both of them are corrupted by randomly and independently flipping negative labels into positive ones with probability $q$ equals $\{0.1, 0.2, 0.4\}$ and $\{0.05, 0.1, 0.2\}$, respectively. All datasets are described in~\Cref{tab.datasets}, including the number of samples, features, classes, and the average size of a candidate set.

\begin{table}[th!]
\centering
\caption{Statistics on artificial, real-world, and vision datasets show significant differences in all properties.}
\label{tab.datasets}
\begin{tabular}{@{}l@{\,}r@{\;\,}r@{\;\,}r@{\;\,}r@{}}
\toprule
\bfseries Dataset & \bfseries Size & \bfseries Dim & $\bm{C}$ & \bfseries $\bm{\mathbb{E}(|S|)}$ \\
\midrule
\multicolumn{5}{c}{Artificial datasets} \\
\cmidrule(l{3pt}r{3pt}){1-5}
Bibtex & 7395 & 1836 & 159 & 1.23 \\
Birds & 351 & 260 & 19 & 1.86 \\
Enron & 1702 & 1001 & 53 & 3.38 \\
Medical & 978 & 1449 & 45 & 1.25 \\
Scene & 2407 & 294 & 6 & 1.07 \\
\cmidrule(l{3pt}r{3pt}){1-5}
\multicolumn{5}{c}{Real-world datasets} \\
\cmidrule(l{3pt}r{3pt}){1-5}
MIRFlickr & 10433 & 100 & 7 & 1.77 \\
Music-emotion & 6833 & 98 & 11 & 2.42 \\
Music-style & 6839 & 98 & 10 & 1.44 \\
YeastBP & 6139 & 6139 & 217 & 2.22 \\
YeastCC & 6139 & 6139 & 50 & 3.40 \\
\cmidrule(l{3pt}r{3pt}){1-5}
\multicolumn{5}{c}{Vision datasets\protect\footnotemark[3]}%
\\
\cmidrule(l{3pt}r{3pt}){1-5}
VOC2007 & 9963 & $224\!\!\times\!\!224\!\!\times\!\!3$ & 20 & 1.56 \\
COCO2014 & 122218 & $224\!\!\times\!\!224\!\!\times\!\!3$ & 80 & 2.92 \\
\bottomrule
\end{tabular}
\end{table}

\paragraph{Baseline methods.}
The proposed method \our{} is compared to seven state-of-the-art PML approaches. These approaches include CPLST~\cite{chen2012feature}, ML-kNN~\cite{zhang2007ml},  PAR-MAP and PAR-VLS (family of PARTICLE~\cite{zhang2020partial}), and PML-NI~\cite{xie2021partial}. CPLST is a label-embedding approach that integrates the concepts of principal component analysis and canonical correlation analysis. ML-kNN is a nearest neighbor-based multi-label classification method. PARTICLE transforms the PML task into a multi-label problem through a label propagation procedure and then induces a calibrated label ranking model. It can generate a multi-label classifier with either virtual label splitting (PAR-VLS) or maximum a posteriori inference (PAR-MAP). PML-NI uses the generation process of noisy labels in the candidate set to solve PML problems. Each of these methods has distinct advantages and disadvantages. For example, CPLST allows for integrating multiple features, while ML-kNN is simpler and easier to use.

Furthermore, for experiments on VOC2007 and COCO2014, we additionally compare \our{} to other state-of-the-art PML approaches: fPML~\cite{yu2018feature} and PML-LRS~\cite{sun2019partial}. Moreover, we also consider three state-of-the-art multi-label classification methods: Binary Cross Entropy (BCE), ASL~\cite{ridnik2021asymmetric} and Query2Label~\cite{liu2021query2label}. Finally, we consider the most recent curriculum disambiguation-based method CDCR~\cite{sun2022deep}. fPML learns a matrix that associates instances with labels based on estimated association confidence. PML-LRS decomposes the label matrix into a ground-truth label matrix and an irrelevant label matrix, with the former constrained to be low-rank and the latter constrained to be sparse. ASL is based on the asymmetric properties of PML classification tasks, operating differently on positive and negative samples. Finally, CDCR uses highly confident prediction to create a curriculum for label disambiguation.

\addtocounter{footnote}{1}%
\footnotetext[3]{The images from the VOC2017 and COCO2014 datasets were rescaled to $224\!\times\!224$ in the RGB color scale to improve the processing performance of the neural network.}%

\paragraph{Setups.}
We conducted experiments in two different setups. The first setup is for relatively small artificial and real-world datasets commonly used with shallow machine-learning baseline methods. Another setup is for larger datasets, typically used with deep networks.

\begin{figure}[t!]
    \centering
    \includegraphics[width=\columnwidth]{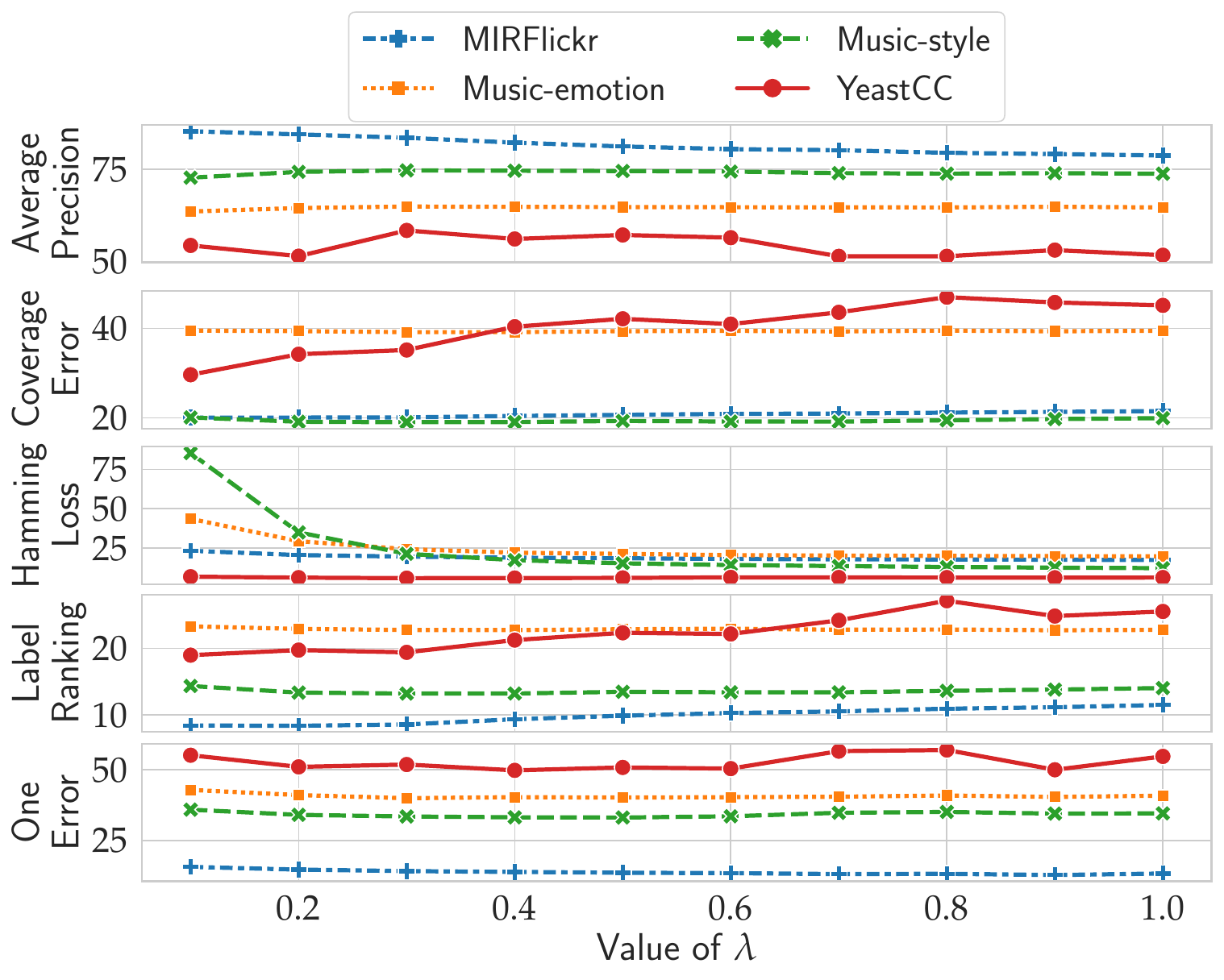}
    \caption{Results for four real-world datasets depending on the value of the $\lambda$ coefficient from the \our{} loss from~\cref{eq:promil}. For average precision, the higher value, the better. For the remaining metrics, the lower value, the better. We observe a strong trend of decreasing Hamming loss for increasing $\lambda$. However, it does not generalize on other metrics, including the most important average precision, which behavior significantly differs between datasets. Nevertheless, most importantly, we observe the stability of the method for various hyperparameter values.}
    \label{fig.lambda_real_worlds}
\end{figure}

\begin{table*}[t]\small
\centering
\caption{Mean results with standard deviations from five \our{} cross-validation runs on the small real-world datasets. For average precision, the higher value, the better. For the remaining metrics, the lower value, the better. The best result is shown in bold, and the second best is in bold italic. Our method has been regarded as the best or second-best result in 22 out of 25 cases. Moreover, it is the best in 13 out of 25 cases, demonstrating its effectiveness and efficiency despite its simplicity.}
\label{tab.real_worlds}
\begin{tabular}{@{}r@{\;\;}l@{\;}r@{\qquad\;}r@{\qquad}r@{\qquad}r@{\qquad}r@{}}
\toprule
 & \bfseries Method & \bfseries Average precision & \bfseries Coverage & \bfseries Hamming loss & \bfseries Label ranking & \bfseries One error \\
\midrule
\multirow[c]{6}{*}{\rotatebox{90}{MIRFlickr}} 
 & CPLST        & 78.67\,(0.53)                   & 22.88\,(0.20)                   & 62.66\,(0.48)                   & 12.62\,(0.85)                  & 29.48\,(2.39)                    \\
 & ML-kNN       & 69.12\,(0.40)                   & 27.07\,(0.22)                   & 76.13\,(0.13)                   & 18.43\,(1.35)                  & 45.17\,(12.32)                   \\
 & PAR-MAP      & \bfseries 86.34\,(0.35)         & \bfseries\itshape 20.22\,(0.09) & \bfseries 15.51\,(0.21)         & \bfseries\itshape 8.78\,(0.10) & \bfseries\itshape 14.30\,(0.73)  \\
 & PAR-VLS      & 69.39\,(0.27)                   & 26.55\,(0.18)                   & 17.34\,(0.16)                   & 20.46\,(0.19)                  & \bfseries 14.00\,(0.45)          \\
 & PML-NI       & 78.98\,(0.38)                   & 22.83\,(0.15)                   & 22.14\,(0.29)                   & 12.26\,(0.22)                  & 29.71\,(1.38)                    \\
 & \our{} (our) & \bfseries\itshape 86.24\,(0.52) & \bfseries 20.18\,(0.37)         & \bfseries\itshape 15.98\,(0.13) & \bfseries 8.70\,(0.42)         & 14.57\,(0.80)                    \\
\cline{1-7}
\multirow[c]{6}{*}{\rotatebox{90}{Music-emotion}} 
 & CPLST        & \bfseries\itshape 61.45\,(0.42) & \bfseries\itshape 40.82\,(0.44)  & 77.86\,(0.06)                   & \bfseries\itshape 24.36\,(0.36) & 46.14\,(0.66)                    \\
 & ML-kNN       & 55.27\,(0.36)                   & 46.86\,(0.33)                    & 85.05\,(0.33)                   & 30.44\,(0.37)                   & 54.65\,(0.75)                    \\
 & PAR-MAP      & 59.12\,(0.50)                   & 42.02\,(0.58)                    & 22.62\,(0.24)                   & 25.84\,(0.50)                   & 51.88\,(0.86)                    \\
 & PAR-VLS      & 61.16\,(0.48)                   & 40.93\,(0.44)                    & \bfseries\itshape 21.25\,(0.06) & 26.08\,(0.36)                   & \bfseries\itshape 42.66\,(1.23)  \\
 & PML-NI       & 60.74\,(0.24)                   & 40.87\,(0.32)                    & 25.63\,(0.78)                   & 24.57\,(0.25)                   & 47.81\,(0.70)                    \\
 & \our{} (our) & \bfseries 64.57\,(0.23)         & \bfseries 39.45\,(0.25)          & \bfseries 20.92\,(0.13)         & \bfseries 22.61\,(0.22)         & \bfseries 39.89\,(1.29)          \\
\cline{1-7}
\multirow[c]{6}{*}{\rotatebox{90}{Music-style}} 
 & CPLST        & 73.23\,(0.55)                   & 20.56\,(0.57)                   & 85.57\,(0.04)                   & 14.53\,(0.57)                   & 34.79\,(0.72)                    \\
 & ML-kNN       & 68.33\,(0.25)                   & 26.32\,(0.54)                   & 85.62\,(0.05)                   & 19.98\,(0.46)                   & 38.18\,(0.29)                    \\
 & PAR-MAP      & 72.27\,(0.53)                   & 20.44\,(0.63)                   & \bfseries 11.87\,(0.21)         & 14.60\,(0.48)                   & 36.73\,(0.68)                    \\
 & PAR-VLS      & 72.07\,(0.37)                   & 20.35\,(0.51)                   & \bfseries\itshape 11.90\,(0.11) & 16.44\,(0.36)                   & 35.82\,(0.85)                    \\
 & PML-NI       & \bfseries\itshape 73.66\,(0.45) & \bfseries\itshape 19.80\,(0.42) & 16.01\,(1.92)                   & \bfseries\itshape 13.83\,(0.46) & \bfseries\itshape 34.65\,(0.71)  \\
 & \our{} (our) & \bfseries 75.79\,(0.47)         & \bfseries 18.25\,(0.48)         & 12.40\,(0.38)                   & \bfseries 12.38\,(0.43)         & \bfseries 31.96\,(0.80)          \\
\cline{1-7}
\multirow[c]{6}{*}{\rotatebox{90}{YeastBP}} 
 & CPLST        & 23.63\,(1.31)                   & 53.30\,(0.99)                   & 52.54\,(0.42)                   & 30.80\,(0.84)                   & 74.96\,(1.65)                    \\
 & ML-kNN       & 17.27\,(1.63)                   & 64.38\,(2.30)                   & 99.90\,(0.02)                   & 35.76\,(1.25)                   & 83.20\,(2.42)                    \\
 & PAR-MAP      & 13.19\,(0.70)                   & 65.44\,(1.14)                   & 4.83\,(0.09)                    & 38.77\,(0.94)                   & 90.63\,(1.35)                    \\
 & PAR-VLS      & 12.67\,(0.42)                   & 72.13\,(0.49)                   & \bfseries 3.70\,(0.04)          & 43.57\,(0.61)                   & 88.56\,(0.59)                    \\
 & PML-NI       & \bfseries 40.06\,(1.21)         & \bfseries 40.91\,(1.05)         & \bfseries\itshape 4.24\,(0.09)  & \bfseries\itshape 21.30\,(0.46) & \bfseries 55.77\,(1.80)          \\
 & \our{} (our) & \bfseries\itshape 32.14\,(1.27) & \bfseries\itshape 42.44\,(1.88) & 5.54\,(0.78)                    & \bfseries 20.55\,(0.77) & \bfseries\itshape 67.44\,(1.40)          \\
\cline{1-7}
\multirow[c]{6}{*}{\rotatebox{90}{YeastCC}} 
 & CPLST        & 41.02\,(1.29)                   & 37.59\,(1.14)                   & 53.48\,(0.52)                  & 26.02\,(0.84)                   & 64.46\,(2.66)                    \\
 & ML-kNN       & 30.42\,(1.76)                   & 52.30\,(1.17)                   & 99.79\,(0.04)                  & 36.37\,(1.40)                   & 75.06\,(2.94)                    \\
 & PAR-MAP      & 21.61\,(0.80)                   & 50.82\,(0.83)                   & 9.41\,(0.35)                   & 38.12\,(0.90)                   & 88.13\,(0.45)                    \\
 & PAR-VLS      & 17.78\,(0.47)                   & 63.10\,(0.41)                   & 8.28\,(0.10)                   & 48.03\,(0.74)                   & 88.74\,(0.51)                    \\
 & PML-NI       & \bfseries 56.25\,(1.30)         & \bfseries\itshape 28.88\,(1.42) & \bfseries 6.64\,(0.19)         & \bfseries 18.24\,(1.16)         & \bfseries 46.84\,(1.79)          \\
 & \our{} (our) & \bfseries\itshape 55.37\,(1.73) & \bfseries 28.04\,(1.74)         & \bfseries\itshape 7.20\,(0.80) & \bfseries\itshape 18.58\,(1.52) & \bfseries\itshape 52.42\,(2.38)  \\
\bottomrule
\end{tabular}
\end{table*}

For smaller datasets, we use cross-validation with 80\% training and 20\% testing sets to demonstrate our results with standard deviations. We followed the guidelines from the original publications and applied the specified hyperparameter where possible. Moreover, we applied an MLP model with two hidden layers to our approach. The training was carried out for 500 epochs with Adam's optimizer and stochastic gradient descent as the optimization method. We chose the parameter $\lambda$ in our cost function \cref{eq:promil} separately for each dataset using the grid search technique and narrowing it to an interval between $0.02$ and $1$ (see~\Cref{fig.lambda_real_worlds}). For large vision experiments, we use the official train-test splits. Following~\cite{ridnik2021asymmetric} as a backbone we employ the TResNet-L architecture~\cite{ridnik2021tresnet}, pretrained on the ImageNet1K dataset~\cite{deng2009imagenet}. All images are resized to $224\times224$. Moreover, as augmentation, we use standard RandAugment~\cite{cubuk2020randaugment} combined with random cutout. Adam optimizer is applied with a one-cycle learning rate policy~\cite{smith2019super}. A bayesian optimization and hyperband~\cite{falkner2018bohb} is used to tune hyper-parameters, finding the optimal learning rate between $2\cdot10^{-5}$ and $2\cdot10^{-4}$, weight decay between $10^{-7}$ and $5\cdot10^{-4}$, and the $\lambda$ for \our{} function between $0.02$ and $1$. The exponential moving average is applied to model parameters with a decay rate of 0.9997.

All experiments were implemented using PyTorch\footnote{\url{https://pytorch.org}} and ran on NVIDIA GeForce RTX $3080$ and Tesla V100.

\section{Results and discussion.}

\begin{figure}[t]
    \centering
    \includegraphics[width=\columnwidth]{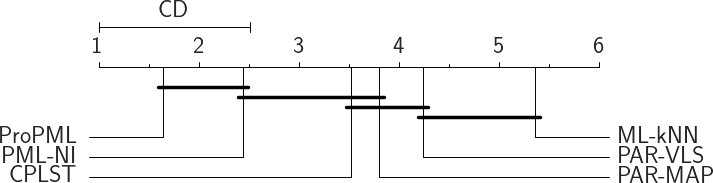}
    \caption{Critical difference diagrams comparing results on small real-world datasets shown in~\Cref{tab.real_worlds} (smaller is better). \our{} performs significantly better than all baseline methods except PML-NI.}
    \label{fig.cd_real_worlds}
\end{figure}

\begin{table*}[th!]\small
\centering
\caption{Mean results with standard deviations from five \our{} cross-validation runs on the small artificial datasets for three false to true label ratios $q=50\%$, $100\%$, or $150\%$. For average precision, the higher value, the better. For the coverage, the lower value, the better. The results of the remaining metrics are provided in the Supplementary Materials. The best result is shown in bold, and the second best is in bold italic. Our method has been regarded as the best or second-best result in 21 out of 30 cases. Moreover, it is the best in 13 out of 30 cases, demonstrating its effectiveness and efficiency despite its simplicity.}
\label{tab.synthetic_part2}
\begin{tabular}{@{}r@{\;}l@{\quad}r@{\;}r@{\;}r@{\quad}r@{\;}r@{\;}r@{}}
\toprule
 & Method & \multicolumn{3}{c}{Average precision} & \multicolumn{3}{c}{Coverage} \\
\midrule
 && $q = 50\%$ & $q = 100\%$ & $q = 150\%$ & $q = 50\%$ & $q = 100\%$ & $q = 150\%$ \\
 \cmidrule(l{2pt}r{10pt}){3-5}
 \cmidrule(l{2pt}r{2pt}){6-8}

\multirow[c]{6}{*}{\rotatebox{90}{Bibtex}} 
 & CPLST           & \bfseries\itshape 54.06\,(0.49) & \bfseries\itshape 49.19\,(0.69) & \bfseries\itshape 48.42\,(0.69) & \bfseries\itshape 18.70\,(0.83) & \bfseries\itshape 17.42\,(0.61) & \bfseries\itshape 16.74\,(0.26)       \\
 & ML-kNN          & 32.04\,(0.68)              & 29.25\,(0.76)           & 28.48\,(0.83)                                & 34.35\,(0.40)           & 34.28\,(0.51)           & 34.61\,(0.45)                                         \\
 & PAR-MAP         & 19.86\,(0.91)              & 18.13\,(0.67)           & 16.97\,(0.42)                                & 46.30\,(0.33)           & 46.93\,(0.09)           & 47.31\,(0.21)                                         \\
 & PAR-VLS         & 25.55\,(0.25)              & 21.28\,(0.43)           & 18.23\,(0.54)                                & 49.92\,(0.35)           & 52.12\,(0.42)           & 52.43\,(0.59)                                         \\
 & PML-NI          & \bfseries 54.45\,(0.52)    & \bfseries 49.54\,(0.65) & \bfseries 48.76\,(0.71)                      & \bfseries 18.00\,(0.83) & \bfseries 16.84\,(0.56) & \bfseries 16.19\,(0.30)                               \\
 & \our{} (our)    & 53.22\,(0.64)              & 48.59\,(0.95)           & 47.46\,(0.64)                                & 23.53\,(0.88)           & 21.65\,(0.54)           & 26.88\,(0.34)                                         \\
\cline{1-8}
\multirow[c]{6}{*}{\rotatebox{90}{Birds}} 
 & CPLST         & 56.25\,(3.67)             & 53.45\,(1.93)            & 52.25\,(1.27)                                & 29.93\,(2.25)           & 30.06\,(1.37)           & 30.25\,(0.94)                    \\
 & ML-kNN        & 55.66\,(4.49)             & 51.00\,(1.91)            & 51.16\,(2.46)                                & 27.22\,(2.11)           & 27.66\,(1.84)           & \bfseries\itshape 27.64\,(2.52)  \\
 & PAR-MAP       & 47.17\,(2.55)             & 43.90\,(2.06)            & 42.96\,(1.69)                                & 34.22\,(2.85)           & 36.39\,(3.44)           & 36.55\,(2.33)                    \\
 & PAR-VLS       & 57.56\,(1.92)             & 53.00\,(3.62)            & 49.44\,(3.66)                                & 32.78\,(1.16)           & 36.82\,(3.54)           & 37.30\,(3.59)                    \\
 & PML-NI        & \bfseries\itshape 63.74\,(1.44) & \bfseries\itshape 58.15\,(1.97) & \bfseries\itshape 57.31\,(2.86) & \bfseries 24.15\,(1.27) & \bfseries\itshape 24.19\,(1.26) & \bfseries 24.57\,(1.23)  \\
 & \our{} (our)  & \bfseries 66.88\,(2.00) & \bfseries 66.77\,(2.64) & \bfseries 62.02\,(3.18)                         & \bfseries\itshape 24.35\,(1.20) & \bfseries 22.50\,(1.85) & 28.13\,(2.46)            \\
\cline{1-8}
\multirow[c]{6}{*}{\rotatebox{90}{Enron}}
 & CPLST          & 52.30\,(0.83) & 50.20\,(0.75) & 47.84\,(0.52)                                                       & 45.46\,(1.19)           & 39.89\,(1.43)          & 39.49\,(0.62)                     \\
 & ML-kNN         & 62.53\,(1.48) & 56.82\,(0.69) & 56.07\,(0.70)                                                       & \bfseries 24.93\,(0.58) & \bfseries 25.58\,(0.75)& \bfseries 26.00\,(0.65)           \\
 & PAR-MAP        & 59.29\,(1.93) & 50.58\,(0.98) & 49.68\,(0.71)                                                       & 28.19\,(1.13)           & \bfseries\itshape 28.03\,(1.07)          & 27.73\,(1.11)   \\
 & PAR-VLS        & 53.32\,(2.11) & 42.54\,(1.80) & 40.49\,(0.87)                                                       & 45.29\,(0.93)           & 41.76\,(1.28)          & 40.67\,(1.88)                     \\
 & PML-NI         & \bfseries\itshape 64.35\,(1.26) & \bfseries\itshape 61.42\,(0.89) & \bfseries\itshape 59.72\,(0.40) & 32.09\,(1.62)           & 28.41\,(1.30)          & \bfseries\itshape 27.37\,(0.66)   \\ 
 & \our{} (our)   & \bfseries 70.10\,(1.01)& \bfseries 66.97\,(0.50) & \bfseries 66.92\,(0.41)                          & \bfseries\itshape 26.63\,(1.11) & 29.59\,(1.66) & 27.69\,(1.02)                      \\
\cline{1-8}
\multirow[c]{6}{*}{\rotatebox{90}{Medical}}
 & CPLST        & 79.57\,(2.55)           & 68.06\,(2.84)           & 67.17\,(2.72)                                   & 8.12\,(1.90)           & 7.86\,(2.47)           & 8.70\,(2.04)                                  \\ 
 & ML-kNN       & 80.78\,(0.65)           & 71.05\,(1.05)           & 70.20\,(1.40)                                   & 5.78\,(0.63)           & 6.53\,(0.64)           & 6.57\,(0.68)                                  \\ 
 & PAR-MAP      & 63.86\,(1.46)           & 59.57\,(1.94)           & 59.34\,(1.44)                                   & 9.96\,(0.48)           & 13.00\,(0.62)          & 12.98\,(0.85)                                 \\ 
 & PAR-VLS      & 78.07\,(1.05)           & 63.06\,(2.05)           & 61.00\,(2.37)                                   & 11.22\,(0.43)          & 16.97\,(0.36)          & 17.14\,(0.36)                                 \\ 
 & PML-NI       & \bfseries 90.84\,(0.52) & \bfseries 82.77\,(1.07) & \bfseries 81.28\,(0.79)                         & \bfseries 2.58\,(0.52) & \bfseries 3.26\,(0.33) & \bfseries 3.57\,(0.36)                        \\ 
 & \our{} (our) & \bfseries\itshape 89.19\,(0.91) & \bfseries\itshape 77.81\,(0.51) & \bfseries\itshape 77.47\,(3.90) & \bfseries\itshape 3.75\,(0.79) & \bfseries\itshape 4.71\,(0.72) & \bfseries\itshape 4.67\,(0.80)\\
\cline{1-8}
\multirow[c]{6}{*}{\rotatebox{90}{Scene}}
 & CPLST            & 82.74\,(2.20)           & 79.88\,(1.66)           & 80.24\,(1.69)                  & 10.44\,(1.73)          & 10.41\,(1.01)          & 10.26\,(1.03)                                          \\
 & ML-kNN           & \bfseries\itshape 86.27\,(1.18) & \bfseries\itshape 84.12\,(1.09) & 83.89\,(1.24)  & \bfseries\itshape 8.10\,(0.52) & \bfseries\itshape 8.95\,(0.62) & 9.04\,(0.74)                           \\
 & PAR-MAP          & 85.06\,(1.02) & 83.87\,(0.81) & \bfseries\itshape 84.09\,(0.78)                    & 8.61\,(0.83)           & 9.08\,(0.53)           & 8.89\,(0.55)                                           \\
 & PAR-VLS          & 84.48\,(0.80) & 83.85\,(0.80) & 84.07\,(0.82)                                      & 8.79\,(0.58)           & \bfseries\itshape 8.95\,(0.55)           & \bfseries\itshape 8.81\,(0.60)       \\
 & PMLNI            & 84.00\,(2.15)           & 80.68\,(2.07)           & 80.84\,(1.77)                  & 9.56\,(1.56)           & 9.88\,(1.08)           & 9.83\,(1.01)                                           \\
 & \our{} (our)     & \bfseries 88.70\,(1.08) & \bfseries 86.78\,(1.17) & \bfseries 86.77\,(1.01)        & \bfseries 7.10\,(0.82) & \bfseries 7.28\,(0.74) & \bfseries 7.21\,(0.64)                                 \\
\bottomrule
\end{tabular}
\end{table*}

\begin{table*}[th!]\small
	\centering
	\caption{Results on VOC2007 dataset for three noise ratios $q\in\{0.1,0.2, 0.4\}$. For all metrics (presented in \%), the higher value, the better. The best result is shown in bold, and the second best is in bold italic. Baseline results are presented as reported in~\cite{sun2022deep}. Our method achieves the top results for the noise ratio $q\in\{0.2, 0.4\}$ and is on par with more complex CDCR for $q=0.1$. Moreover, it is the least affected by the increase in noise ratio.}   
	\label{table:voc_comparison_result}
	\begin{tabular}{l rrr@{\qquad}rrr@{\qquad}rrr}
		\toprule
		\multirow{2}*{Method} &
		\multicolumn{3}{c}{$q=0.1$} &  \multicolumn{3}{c}{$q=0.2$} & \multicolumn{3}{c}{$q=0.4$} \\
		\cmidrule(l{2pt}r{18pt}){2-4} \cmidrule(l{2pt}r{18pt}){5-7} \cmidrule(l{2pt}r{2pt}){8-10}
		& mAP & CF1 & OF1 & mAP & CF1 & OF1 & mAP & CF1 & OF1\\ 
        \midrule
		PML-NI & 80.32 &  65.64 &  68.56 & 77.63 &  63.44 &  67.26 & 69.95 &  58.21 &  63.39\\
		fPML  & 70.36 &  61.49 &  66.01 & 75.29 &  64.41 &  66.83 & 63.22 &  55.20 &  60.70\\
		PMLRS & 80.68 &  66.10 &  68.72 & 77.08 &  62.99 &  66.96 & 63.86 &  55.45 &  61.01\\
		PAR-MAP &  69.61 &  66.69 &  67.12 & 67.66 &  64.43 &  66.38 & 65.69 &  61.18 &  64.45 \\
		PAR-VLS & 72.50 &  64.64 &  65.40 & 70.54 &  63.56 &  65.12 & 69.25 &  62.05 &  64.28\\
		\midrule
		BCE  & 84.04 &  68.46 &  70.09 & 80.43 &  63.86 &  68.63 & 71.35 &  59.95 &  65.05\\
		ASL & 86.62 &  68.50 &  72.29 & 83.89 &  67.34 &  71.07 &   74.99 &  62.54 &  67.52\\
		Query2Label & 87.14 &  68.10 &  72.82 & 84.81 &  66.75 &  71.72 & 79.90 &  63.88 &  68.84\\
		CDCR  & \textbf{90.12} &  \bfseries\itshape{71.07} &  \bfseries\itshape{72.96} & \bfseries\itshape{88.94} & \bfseries\itshape{70.99} &  \bfseries\itshape{73.00} & \bfseries\itshape{86.15} &  \bfseries\itshape{70.07} &  \bfseries\itshape{71.94} \\
		\midrule
		\our{} (Our)  & \bfseries\itshape {89.43} &  \textbf{71.39} & \textbf{73.10} & \textbf{88.96} &  \textbf{72.89} &  \textbf{73.40} & \textbf{86.82} &  \textbf{72.36} &  \textbf{72.72} \\
		\midrule
		Supervised ($q=0$) & 90.93 & 78.77 & 77.06 & -- & -- & -- & -- & -- & --\\
		\bottomrule
	\end{tabular}
\end{table*}

\begin{table*}[th!]\small
	\centering
	\caption{Results on COCO2014 dataset for three noise ratios $q\in\{0.1,0.2, 0.4\}$. For all metrics (presented in \%), the higher value, the better. The best result is shown in bold, and the second best is in bold italic. Baseline results are presented as reported in~\cite{sun2022deep}. Our method is on par with more complex CDCR for $q=0.05,0.1$ and achieves slightly worse scores for $q=0.2$. However, it is still better than other loss-based methods, such a ASL.}  
	\label{table:COCO_comparison_result}
	\begin{tabular}{l rrr@{\qquad}rrr@{\qquad}rrr}
		\toprule
		\multirow{2}*{Method} &
		\multicolumn{3}{c}{$q=0.05$} &  \multicolumn{3}{c}{$q=0.1$} & \multicolumn{3}{c}{$q=0.2$} \\
		\cmidrule(l{2pt}r{18pt}){2-4} \cmidrule(l{2pt}r{18pt}){5-7} \cmidrule(l{2pt}r{2pt}){8-10}
		& mAP & CF1 & OF1 & mAP & CF1 & OF1 & mAP & CF1 & OF1\\ 
        \midrule
		PML-NI & 60.36 &  52.97 &  58.70 & 59.87 &  52.61 &  58.56 & 58.92 &  51.81 &  58.20 \\
		fPML  & 57.93 &  51.29 &  57.55 & 57.75 &  51.08 &  57.54 & 57.14 &  50.86 &  57.47 \\
		\midrule
		BCE  & 72.39  & 68.17  & 72.47 & 70.43 &  64.41 &  69.30 & 67.30 &  48.80 &  56.02 \\
		ASL & 76.92  & 61.11  & 66.98 & 75.14 &  60.00 &  66.26 & 72.28 &  57.65 &  64.88 \\
		Query2Label & 76.56 &  60.83 &  66.69 &  75.39 &  60.50 &  66.30 &73.59 &  59.36 &  65.54 \\
		CDCR  & \textbf{77.95} &  \textbf{73.48} &  \textbf{77.42} &  \textbf{77.14} &  \textbf{72.99} &  \textbf{77.00} &  \textbf{76.13} &  \textbf{72.58} &  \textbf{76.58}  \\
        \midrule
        \our{} (Our) & \bfseries\itshape77.15 & \bfseries\itshape71.96 & \bfseries\itshape76.64 & \bfseries\itshape76.03 & \bfseries\itshape71.53 & \bfseries\itshape75.57 & \bfseries\itshape74.92 & \bfseries\itshape62.86 & \bfseries\itshape67.01 \\
		\midrule
		Supervised ($q=0$) & 80.23 & 75.03 & 77.75 & -- & -- & -- & -- & -- & -- \\
		\bottomrule
	\end{tabular}
\end{table*}

In this section, we compare the results of our method with those of baseline approaches in two different setups of experiments for three types of datasets. We investigate how our approach fares against existing techniques and how the results vary in different scenarios.

\paragraph{The real-world datasets.} \Cref{tab.real_worlds} shows the results of the experiment conducted on real-world data. Five-fold cross-validation was used to measure the mean and variance for each method and dataset. The two highest values for each dataset and measure are highlighted, with the best result shown in bold and the second best result presented in bold italic. Our method has been regarded as the best or second-best result in 22 out of 25 cases. Moreover, it is the best in 13 out of 25 cases, with up to $3.12$ gap to the second-best approach in the case of average precision and up to $1.55$ gap in the case of coverage. This trend is also confirmed by the Critical Difference (CD) diagrams~\cite{demvsar2006statistical} in~\Cref{fig.cd_real_worlds}, generated using all the results shown in~\Cref{tab.real_worlds}. They show that for small artificial datasets, \our{} performs significantly better than all baseline methods except PML-NI. While in the case of large datasets PML-NI method produces much worse results than our (see ~\Cref{table:voc_comparison_result,table:COCO_comparison_result}).

In \Cref{fig.lambda_real_worlds}, we observe how the results change depending on the $\lambda$ parameter. There is a clear impact of this parameter on the Hamming loss. It is expected because Hammin loss is the fraction of incorrectly predicted labels. Therefore, it is calculated by summing up the number of false labels predicted as true and dividing it by the total number of labels. At the same time, the second part of the \our{} loss is also responsible for penalizing incorrectly classified false labels. Moreover, $\lambda$ is a coefficient of the second component of the \our{} loss. Therefore, its increase is expected to decrease the second component and, in consequence, the Hamming loss. Unfortunately, similar trends are not observed for other metrics. Therefore, the best solution requires searching for the optimal $\lambda$. Nevertheless, most importantly, we observe the stability of the method for various hyperparameter values.

\begin{figure}[!ht]
     \centering
     \begin{subfigure}[b]{\columnwidth}
         \centering
         \includegraphics[width=\textwidth]{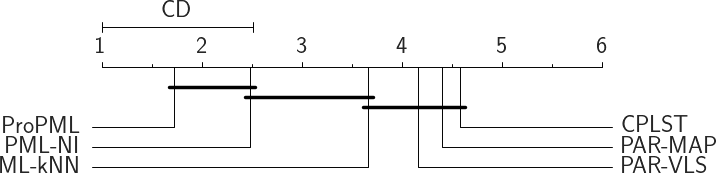}
         \caption{False to true label ratio $q=50\%$.}
         \label{fig.synthetic_50}
     \end{subfigure}
     \hfill
     \begin{subfigure}[b]{\columnwidth}
         \centering
         \includegraphics[width=\textwidth]{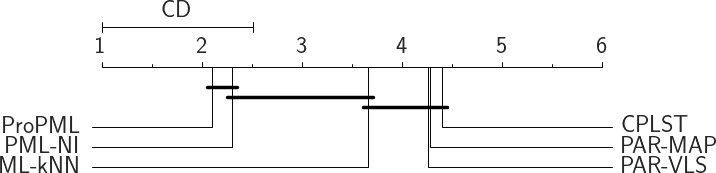}
         \caption{False to true label ratio $q=100\%$.}
         \label{fig.synthetic_100}
     \end{subfigure}
     \hfill
     \begin{subfigure}[b]{\columnwidth}
         \centering
         \includegraphics[width=\textwidth]{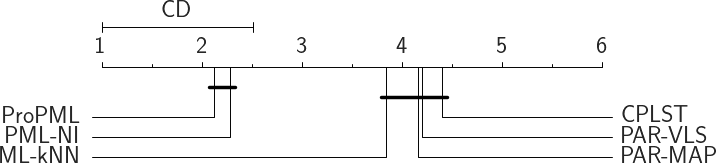}
         \caption{False to true label ratio $q=150\%$.}
         \label{fig.synthetic_150}
     \end{subfigure}
        \caption{Critical difference diagrams comparing results on small artificial datasets with false to true label ratio $q=50\%$, $100\%$, or $150\%$ shown in~\Cref{tab.synthetic_part2} and in the Supplementary Materials (smaller is better). \our{} performs significantly better than all baseline methods except PML-NI. Moreover, the gap between those two models and the remaining approaches grows together with increased $q$.}
        \label{fig.pml_synthetic}
\end{figure}

\paragraph{Artificial datasets.} In~\Cref{tab.synthetic_part2} and the Supplementary Materials, we show results for synthetic datasets created by adding noisy labels with false to true labels ratio $q=50\%$, 100\%, and 150\%. These results confirm the effectiveness of our method against the considered approaches, with up to $7.2$ gap to the second-best approach in the case of average precision and up to $1.6$ gap in the case of coverage. Similar to the real-world datasets, this trend is also confirmed by the Critical Difference (CD) diagrams in~\Cref{fig.pml_synthetic}, generated using all the results shown in~\Cref{tab.synthetic_part2} and the Supplementary Materials. Moreover, again, \our{} performs significantly better than all baseline methods except PML-NI. However, interestingly, the gap between those two models and the remaining approaches grows together with increased $q$. Please, consult the Supplementary Materials to analyze the influence of parameter $\lambda$ from \cref{eq:promil} on the considered metrics.

\paragraph{Vision datasets.} \Cref{table:voc_comparison_result} shows the results of our experiments on the VOC2007 dataset. As presented, our method outperforms all state-of-the-art PML approaches, as well as methods dedicated to multi-label image classification. The performance is on par with more complex, curriculum disambiguation-based CDCR, providing slightly better results in higher noise conditions, with the improvement of $0.67$ percentage points in mean Average Precision (mAP) and $2.29$ percentage points Class average F1 score (CF1) for $q=0.4$. Results for COCO2014 are presented in \Cref{table:COCO_comparison_result}. Again our method outperforms standard PML and multi-label image classification methods. However, contrary to VOC2007, in the case of COCO2014, \our{} falls slightly behind the CDCR approach. We hypothesize that this is caused by a much larger number of possible labels (80 in COCO2014 vs. 20 in VOC2007) and a larger variance in semantic characteristics between classes making this scenario much more imbalanced. However, note that \our{} is much more straightforward than CDCR as it does not require an external curriculum-building process nor consistency regularization and can be used as a drop-in replacement for regular binary cross-entropy loss.

\section{Conclusions}

In this paper, we introduced \our{}, a direct and easy-to-implement probabilistic approach to partial multi-label learning that adapts binary cross entropy to PML setup. It considers the expected value of positively predicted labels from a candidate set and uses it to encourage the model to find the true labels. As a result, it commonly overpasses the existing solutions, especially for the high number of false labels in a candidate set, as we present in multiple experiments.

In the future, we want to focus on adopting our function to other target tasks, such as detection or segmentation, and other modalities, such as tabular data, text, sound, or graphs.



\section*{Acknowledgements}

This research was partially funded by the National Science Centre, Poland, grants no.  2020/39/D/ST6/01332 (work by \L{}ukasz Struski), 2022/47/B/ST6/03397 (work by Bartosz Zieli\'nski), and 2021/41/B/ST6/01370 (work by Adam Pardyl and Jacek Tabor).
Some experiments were performed on servers purchased with funds from a grant from the Priority Research Area (Artificial Intelligence Computing Center Core Facility) under the Strategic Programme Excellence Initiative at Jagiellonian University. We gratefully acknowledge Polish high-performance computing infrastructure PLGrid (HPC Centers: ACK Cyfronet AGH) for providing computer facilities and support within computational grant no. PLG/2022/015753.


\bibliographystyle{IEEEtran}
\bibliography{DSAA23/conference}
\end{document}

%% file: voc_plot.tex
\begin{figure}[t]
\begin{center}
\begin{small}
\begin{sc}
\begin{tikzpicture}
\tikzstyle{every node}=[font=\scriptsize]
\begin{axis}[
    title={VOC2007},
    xlabel={Noise ratio},
    ylabel={mAP (\%)},
    xtick={0.1, 0.2, 0.4},
    ymajorgrids=true,
    grid style=dashed,
    scale only axis,
    height=5cm,
    width=.42\textwidth,
    legend pos=south west,
    every axis plot/.append style={thick},
    smooth,
    legend cell align={left}
]
\addplot[color=violet, mark=square*]coordinates {(0.1,84.04)(0.2,80.43)(0.4,71.35)};
\addlegendentry{BCE}
\addplot[color=teal, mark=pentagon*]coordinates {(0.1,86.62)(0.2,83.89)(0.4,74.99)};
\addlegendentry{ASL}
\addplot[color=green, mark=diamond*]coordinates {(0.1,87.14)(0.2,84.81)(0.4,79.90)};
\addlegendentry{Query2Label}
\addplot[color=red, mark=triangle*]coordinates {(0.1,90.12)(0.2,88.94)(0.4,86.15)};
\addlegendentry{CDCR}
\addplot[color=blue, mark=oplus*]coordinates {(0.1,89.43)(0.2,88.96)(0.4,86.82)};
\addlegendentry{\our{} (our)}

\end{axis}
\end{tikzpicture}
\end{sc}
\end{small}
\end{center}

\caption{The mean Average Precision (mAP) obtained for the top-5 methods on the VOC2007 dataset artificially corrupted to PML by flipping negative labels into positive with probability $q$ (noise ratio) equals $\{0.1, 0.2, 0.4\}$. \our{} obtains outperforms existing methods for higher noise ratio. Moreover, compared to the second-best complex CDCR based on curriculum learning, ProPML requires only loss function modification.}
\label{fig:voc_plot}
\end{figure}